\documentclass[10pt, conference, compsocconf]{IEEEtran}

\usepackage{cite}
\ifCLASSINFOpdf
  \usepackage[pdftex]{graphicx}
\else
\fi
\usepackage[cmex10]{amsmath}
\usepackage{array}
\usepackage{caption}
\usepackage{url}
\usepackage{multirow}
\usepackage{float}

\begin{document}

\title{Nature vs. Nurture: The Role of Environmental Resources in Evolutionary Deep Intelligence}

\author{\IEEEauthorblockN{Audrey G. Chung, Paul Fieguth, and Alexander Wong}
\IEEEauthorblockA{Vision \& Image Processing Research Group\\
Dept. of Systems Design Engineering, University of Waterloo\\
Waterloo, ON, Canada, N2L 3G1\\
Emails: \{agchung, pfieguth, a28wong\}@uwaterloo.ca}
}

\maketitle

\begin{abstract}
\textit{Evolutionary deep intelligence} synthesizes highly efficient deep neural networks architectures over successive generations. Inspired by the nature versus nurture debate, we propose a study to examine the role of external factors on the network synthesis process by varying the availability of simulated environmental resources. Experimental results were obtained for networks synthesized via asexual evolutionary synthesis (1-parent) and sexual evolutionary synthesis (2-parent, 3-parent, and 5-parent) using a 10\% subset of the MNIST dataset. Results show that a lower environmental factor model resulted in a more gradual loss in performance accuracy and decrease in storage size. This potentially allows significantly reduced storage size with minimal to no drop in performance accuracy, and the best networks were synthesized using the lowest environmental factor models.


\end{abstract}


\IEEEpeerreviewmaketitle

\section{Introduction}
\label{Introduction}
In recent years, deep neural networks~\cite{Bengio2009, Graves2013, LeCun2015, Tompson2014} have experienced an explosion in popularity due to their ability to accurately represent complex data and their improved performance over other state-of-the-art machine learning methods. This notable increase in performance has mainly been attributed to increasingly large deep neural network model sizes and expansive training datasets, resulting in growing computational and memory requirements~\cite{Han2015}.

For many practical scenarios, however, these computational requirements make powerful deep neural networks infeasible. Applications such as self-driving cars and consumer electronics are often limited to low-power embedded GPUs or CPUs, making compact and highly efficient deep neural networks very desirable. While methods for directly compressing large neural network models into smaller representations have been developed~\cite{Han2015,LeCun1989,Gong2014,Han2015_2,Chen2015,Sun2016}, Shafiee~\textit{et al.}~\cite{Shafiee2016} proposed a radically novel approach: \textit{Can deep neural networks naturally evolve to be highly efficient?} 

Taking inspiration from nature, Shafiee~\textit{et al.} introduced the concept of \textit{evolutionary deep intelligence} to organically synthesize increasingly efficient and compact deep neural networks over successive generations. Evolutionary deep intelligence mimics biological evolutionary mechanisms found in nature using three computational constructs: i) heredity, ii) natural selection, and iii) random mutation.

While the idea of utilizing evolutionary techniques to generate and train neural networks has previously been explored~\cite{Angeline1994,Gauci2007,Stanley2005,Stanley2002,Tirumala2016}, 
evolutionary deep intelligence~\cite{Shafiee2016} introduces some key differences. In particular, where past works use classical evolutionary computation methods such as genetic algorithms, \cite{Shafiee2016} introduced a novel probabilistic framework that models network genetic encoding and external environmental conditions via probability distributions. Additionally, these previous studies have been primarily focused on improving a deep neural network's performance accuracy, while evolutionary deep intelligence shifts part of the focus to synthesizing highly efficient neural network architectures while maintaining high performance accuracy.

Following the seminal evolutionary deep intelligence paper~\cite{Shafiee2016}, Shafiee~\textit{et al.}~\cite{Shafiee2016_2} further proposed a detailed extension of the original approach via synaptic cluster-driven genetic encoding. This work introduced a multi-factor synapse probability model, and comprehensive experimental results using four well-known deep neural networks~\cite{Shafiee2017} produced significantly more efficient network architectures specifically tailored for GPU-accelerated applications while maintaining state-of-the-art performance accuracy. Additional research has since been conducted to include a synaptic precision constraint during the evolutionary synthesis of network architectures~\cite{Shafiee2017_Restrictions} and to introduce the concept of trans-generational genetic transmission of environmental information~\cite{Shafiee2017_Transgenerational}.

\begin{figure*}[t]
	\centering
	\includegraphics[width=\textwidth]{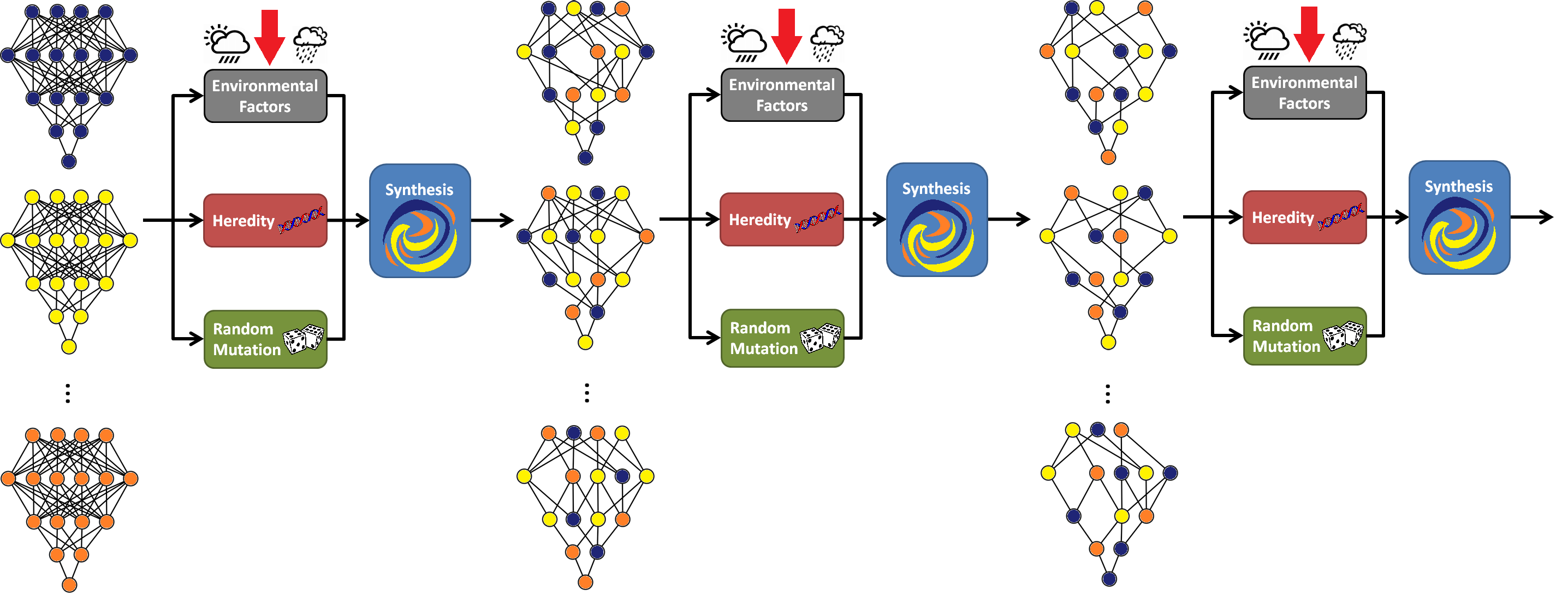}
	\caption{A visualization of the $m$-parent evolutionary synthesis process over successive generations. In this study, we examine the role of simulated external environmental factors (indicated by red arrows) during the $m$-parent network synthesis process over successive generations.}
	\label{fig_AlgDesign}
\end{figure*}

Shafiee~\textit{et al.}'s work~\cite{Shafiee2016,Shafiee2016_2}, however, formulated the evolutionary synthesis process based on asexual reproduction. Recently, Chung~\textit{et al.}~\cite{Chung2017_Mating,Chung2017_Polyploidism} extended the asexual evolutionary sythensis approach used previously in~\cite{Shafiee2016,Shafiee2016_2} to $m$-parent sexual evolutionary synthesis, demonstrating that increasing the number of parent networks resulted in synthesizing networks with improved architectural efficiency with only a $2 - 3\%$ drop in performance accuracy.

In addition to demonstrating state-of-the-art performance in classic computer vision tasks, evolutionary deep intelligence has proven promising in a variety of applications such as real-time motion detection in videos on embedded systems~\cite{Shafiee2017_Embedded}, imaging-based lung cancer detection via an evolutionary deep radiomic sequencer discovery approach using clinical lung CT images with pathologically proven diagnostic data~\cite{Shafiee2017_Radiomics}, and synthesizing even smaller deep neural architectures based on the more recent SqueezeNet v1.1 macroarchitecture for applications with fewer target classes~\cite{Shafiee2017_SquishedNets}. Evolutionary deep intelligence enables the use of powerful deep neural networks in applications where limited computational power, limited memory resources, and/or privacy are of concern.

The effect of environmental factors (e.g., abundance or scarcity of resources) during the evolutionary synthesis process on generations of synthesized network architectures has largely been unexplored. In this study, we aim to better understand the role these environmental resources models play by varying the availability of simulated environmental resources in $m$-parent evolutionary synthesis. The evolutionary deep intelligence method and the proposed environmental resource study are described in Section~\ref{Methods}. The experimental setup and results are presented in Section~\ref{Results}. Lastly, conclusions and future work are discussed in Section~\ref{Conclusion}.

\section{Methods}
\label{Methods}
We propose a study to determine the role of simulated environmental resources during the evolutionary synthesis process (Figure~\ref{fig_AlgDesign}) and their effects on generations of synthesized neural networks. In this work, we leverage the evolutionary deep intelligence framework (as formulated below), using previously proposed cluster-driven genetic encoding~\cite{Shafiee2016_2} and $m$-parent evolutionary synthesis~\cite{Chung2017_Polyploidism} methods, and vary the availability of simulated external environmental resources.

\subsection{Sexual Evolutionary Synthesis}
Let the network architecture be formulated as $\mathcal{H}(N,S)$, where $N$ denotes the set of possible neurons and $S$ the set of possible synapses in the network. Each neuron $n_j \in N$ is connected to neuron $n_k \in N$ via a set of synapses $\bar{s} \subset S$, such that the synaptic connectivity $s_j \in S$ has an associated $w_j \in W$ to denote the connection's strength. In Shafiee~\textit{et al.}'s seminal paper on evolutionary deep intelligence~\cite{Shafiee2016}, the synthesis probability $P(\mathcal{H}_{g}|\mathcal{H}_{g-1}, \mathcal{R}_{g})$ of a synthesized network at generation $g$ is approximated by the synaptic probability $P(S_g|W_{g-1}, R_{g})$; this emulates heredity through the generations of networks, and is also conditional on an environmental factor model $\mathcal{R}_g$ to imitate the availability of resources in the network's environment. Thus, the synthesis probability can be modelled as follows:
\begin{align}
	P(\mathcal{H}_{g}|\mathcal{H}_{g-1}, \mathcal{R}_{g}) \simeq P(S_g | W_{g-1}, R_{g}).
\end{align}

Introducing the synaptic cluster-driven genetic encoding approach, Shafiee \textit{et al.}~\cite{Shafiee2016_2} proposed that the synthesis probability incorporate a multi-factor synaptic probability model and different quantitative environmental factor models at the synapse and cluster levels:

\begin{align}
	P(\mathcal{H}_{g}|&\mathcal{H}_{g-1}, \mathcal{R}_g) = \nonumber \\
	& \prod_{c \in C} \Big[ P(s_{g,c}|W_{g-1}, \mathcal{R}_{g}^c) \cdot \prod_{j \in c} P(s_{g,j}|w_{g-1,j}, \mathcal{R}_g^s) \Big]
\end{align}
where $\mathcal{R}_{g}^c$ and $\mathcal{R}_{g}^s$ represent the environmental factor models enforced during synthesis at the cluster level and the synapse level, respectively. $P(s_{g,c}|W_{g-1}, \mathcal{R}_{g}^c)$ represents the probability of synthesis for a given cluster of synapses $s_{g,c}$. Thus, $P(s_{g,c}|W_{g-1}, \mathcal{R}_{g}^c)$ denotes the likelihood that a synaptic cluster $s_{g,c}$ (for all clusters $c \in C$) will exist in the network architecture in generation $g$ given the cluster's synaptic strength in generation $g-1$ and the cluster-level environmental factor model. Comparably, $P(s_{g,j}|w_{g-1,j}, \mathcal{R}_g^s)$ represents the likelihood of the existence of synapse $j$ within the synaptic cluster $c$ in generation $g$ given the synaptic strength in the previous generation $g-1$ and synapse-level environmental factor model. This multi-factor probability model encourages both the persistence of strong synaptic clusters and the persistence of strong synaptic connectivity over successive generations~\cite{Shafiee2016_2}.

Chung~\textit{et al.}~\cite{Chung2017_Polyploidism} proposed a further modification of the synthesis probability $P(\mathcal{H}_{g}|\mathcal{H}_{g-1}, \mathcal{R}_g)$ via the incorporation of a $m$-parent synthesis process to drive network diversity and adaptability by mimicking sexual reproduction. The synthesis probability was reformulated to combine the cluster and synapse probabilities of $m$ parent networks via some cluster-level mating function $\mathcal{M}_c(\cdot)$ and some synapse-level mating function $\mathcal{M}_s(\cdot)$:

\begin{align}
P(\mathcal{H}_{g(i)}|&\mathcal{H}_{G_i}, \mathcal{R}_{g(i)}) =  \nonumber \\
& \prod_{c \in C} \Big[ P(s_{g(i),c}|\mathcal{M}_c(W_{\mathcal{H}_{G_i}}), \mathcal{R}_{g(i)}^c) \cdot \nonumber \\
& \prod_{j \in c} P(s_{g(i),j}|\mathcal{M}_s(w_{\mathcal{H}_{G_i},j}), \mathcal{R}_{g(i)}^s) \Big].
\end{align} 

\subsection{Mating Rituals of Deep Neural Networks}
In the context of this study, we restrict $\mathcal{H}_{G_i}$ to networks in the immediately preceding generation, i.e., for a newly synthesized network $\mathcal{H}_{g(i)}$ at generation $g(i)$, the $m$ parent networks in $\mathcal{H}_{G_i}$ are from generation $g(i) - 1$. As in~\cite{Chung2017_Polyploidism}, the mating functions are:
\begin{align}
	\mathcal{M}_c(W_{\mathcal{H}_{G_i}}) &= \sum_{k = 1}^{m} \alpha_{c,k} W_{\mathcal{H}_k} \\
	\mathcal{M}_s(w_{\mathcal{H}_{G_i},j}) &= \sum_{k = 1}^{m}\alpha_{s,k} w_{\mathcal{H}_k,j}
\end{align}
where $W_{\mathcal{H}_k}$ represents the cluster's synaptic strength for the $k^{th}$ parent network $\mathcal{H}_k \in \mathcal{H}_{G_i}$. Similarly, $w_{\mathcal{H}_k,j}$ represents the synaptic strength of a synapse $j$ within cluster $c$ for the $k^{th}$ parent network $\mathcal{H}_k \in \mathcal{H}_{G_i}$.

\subsection{The Role of Environmental Resources}
In this work, we aim to study the effects of the cluster-level environmental factor model $\mathcal{R}_{g(i)}^c$. At the cluster level, the existence of all clusters $c \in C$ is determined as:
\begin{align}
	1^c_{g(i)} =
	\begin{cases}
	1 & \mathcal{R}_{g(i)}^c \cdot (1 - |W_{\mathcal{H}_{G_i}}|) \leq \gamma\\
	0 & \text{otherwise}
	\end{cases}
\end{align}
where $1^c_{g(i)}$ incorporates both the strength of the synapses in a cluster ($W_{\mathcal{H}_{G_i}}$) and the cluster-level environmental resources available ($\mathcal{R}_{g(i)}^c$), and $\gamma$ is a randomly generated number between 0 and 1.

Previous studies in evolutionary deep intelligence~\cite{Shafiee2016,Shafiee2016_2,Shafiee2017,Chung2017_Mating,Chung2017_Polyploidism} generally employed a environmental factor model of $\mathcal{R}_{g(i)}^c = 70$, i.e., the probability of existing clusters being stochastically dropped during the evolutionary synthesis process was scaled by 70\%. To better understand the role environmental resources play, we propose a study varying the cluster-level environmental factor model $\mathcal{R}_{g(i)}^c$.

\section{Results}
\label{Results}
\subsection{Experimental Setup}
\begin{figure}[b]
	\centering
	\includegraphics[width=\linewidth]{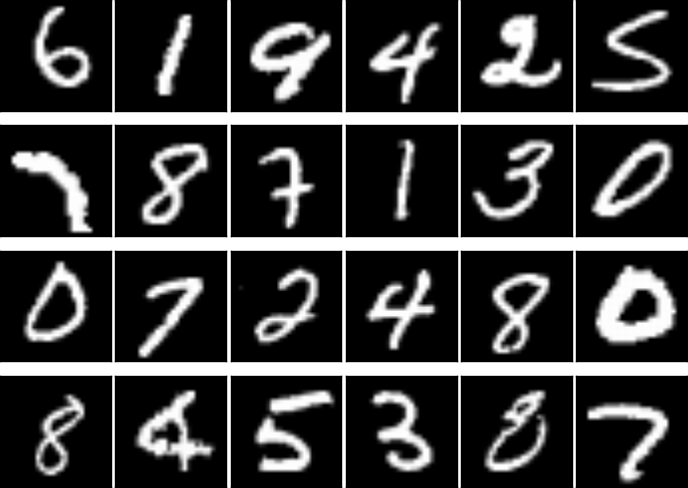}
	\caption{Sample images from the MNIST hand-written digits dataset~\cite{LeCun1998}.}
	\label{fig_MNIST}
\end{figure}

In this study, the $m$-parent evolutionary synthesis of deep neural networks were performed over multiple generations, and the effects of various environmental factor models were explored using 10\% of the MNIST~\cite{LeCun1998} hand-written digits dataset with the first generation ancestor networks trained using the LeNet-5 architecture~\cite{LeCun1998_LeNet}. Figure~\ref{fig_MNIST} shows sample images from the MNIST dataset.

Similar to Shafiee \textit{et al.}'s work~\cite{Shafiee2016_2}, each filter (i.e., collection of kernels) was considered as a synaptic cluster in the multi-factor synapse probability model. In this work, we assessed the synthesized networks using performance accuracy on the MNIST dataset and storage size (representative of the architectural efficiency of a network) of the networks with respect to the computational time required.

To investigate the effects the cluster-level environmental factor model $\mathcal{R}^c_{g(i)}$, we vary $\mathcal{R}^c_{g(i)}$ from 50\% to 95\% at 5\% increments, i.e.,:
\begin{align}
	\mathcal{R}^c_{g(i)} = \{50, 55, 60, 65, 70, 75, 80, 85, 90, 95\}
\end{align}
As in~\cite{Shafiee2016_2}, we use a synapse-level environmental factor model of $\mathcal{R}^s_{g(i)} = 70\%$ to allow for increasingly more compact and efficient network architectures in the successive generations while minimizing any loss in accuracy.

\begin{figure*}
\centering
\begin{tabular}{cc}
\includegraphics[width=.48\textwidth]{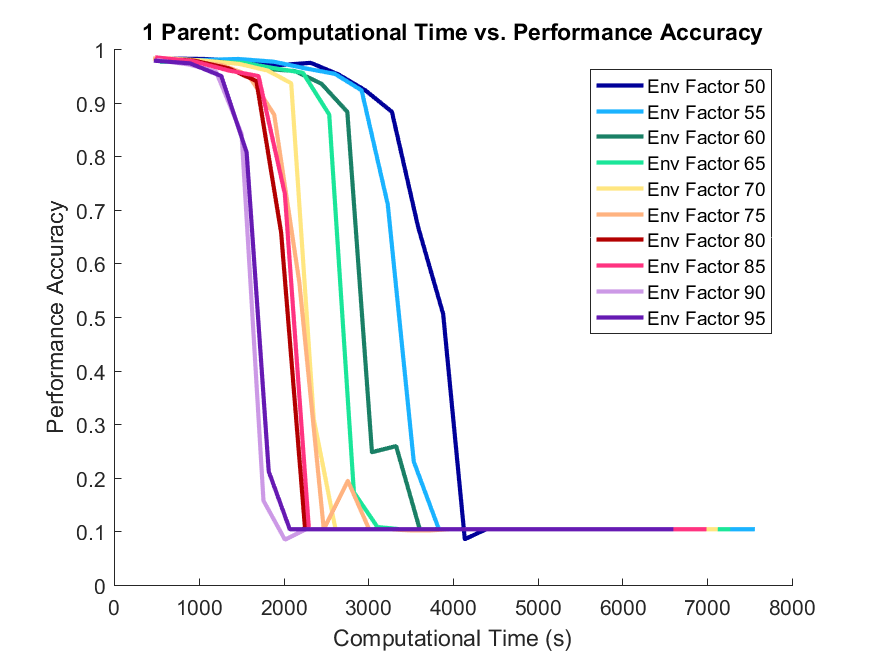}&
\includegraphics[width=.48\textwidth]{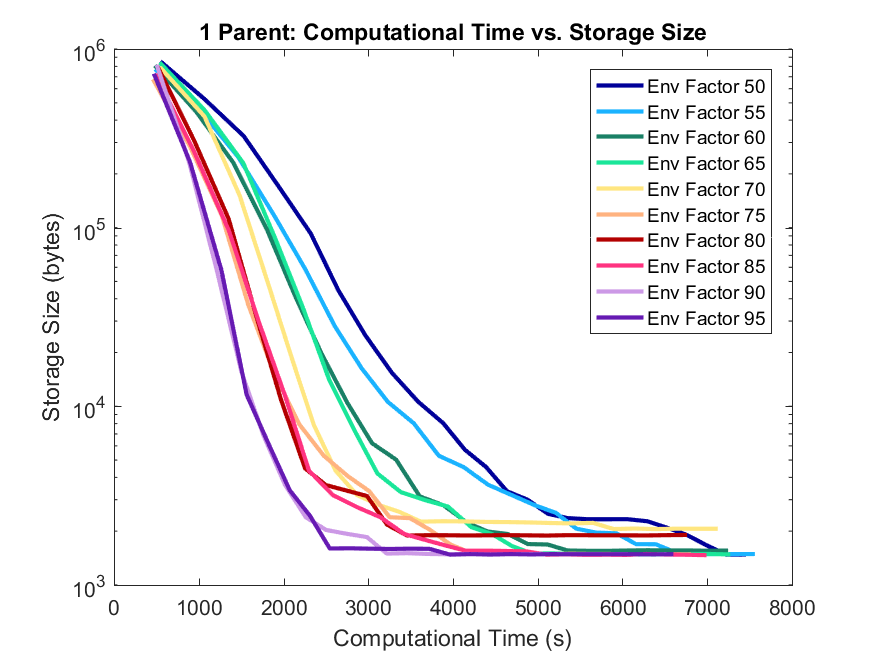}\\\\
\end{tabular}
\caption{Performance accuracy (left) and storage size (right) with respect to computational time for 1-parent evolutionary synthesis using various environmental factor models. Networks synthesized via 1-parent evolutionary synthesis and subject to various environmental factors show a clear monotonic trend.}
\label{fig_Results1}
\end{figure*}

\subsection{Experimental Results}
\begin{figure*}
\centering
\begin{tabular}{cc}
\includegraphics[width=.45\textwidth]{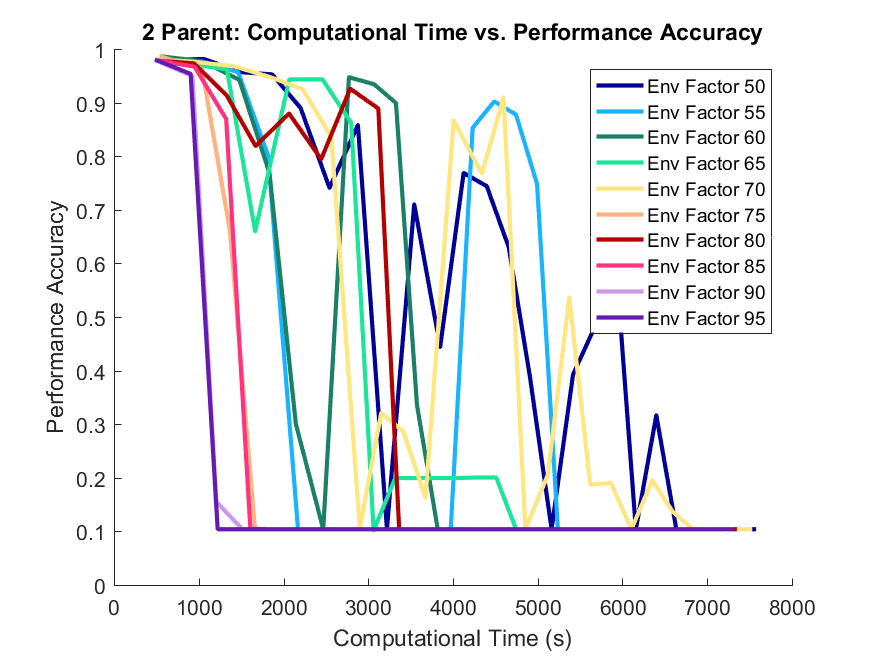}&
\includegraphics[width=.45\textwidth]{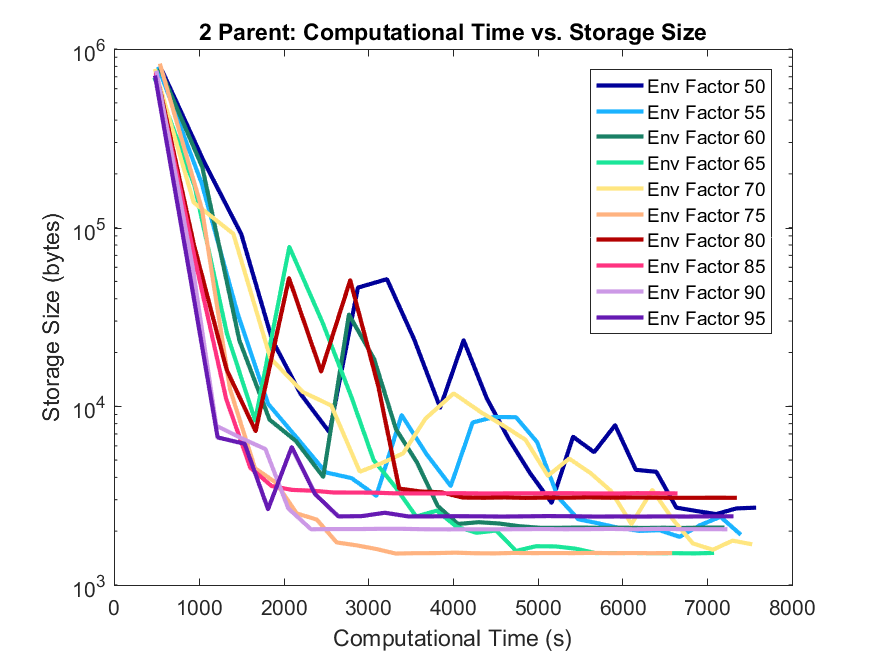}\\\\
\includegraphics[width=.45\textwidth]{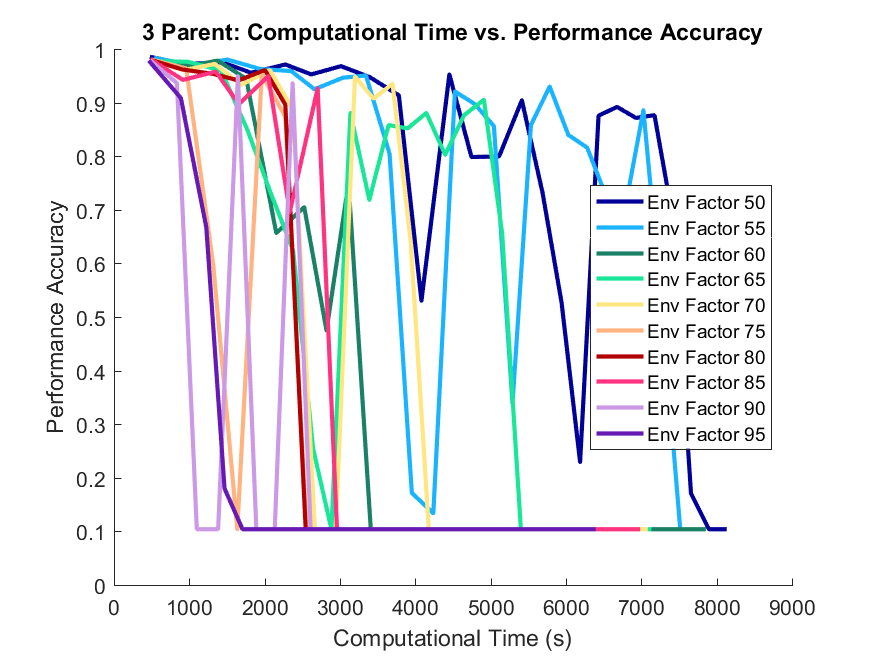}&
\includegraphics[width=.45\textwidth]{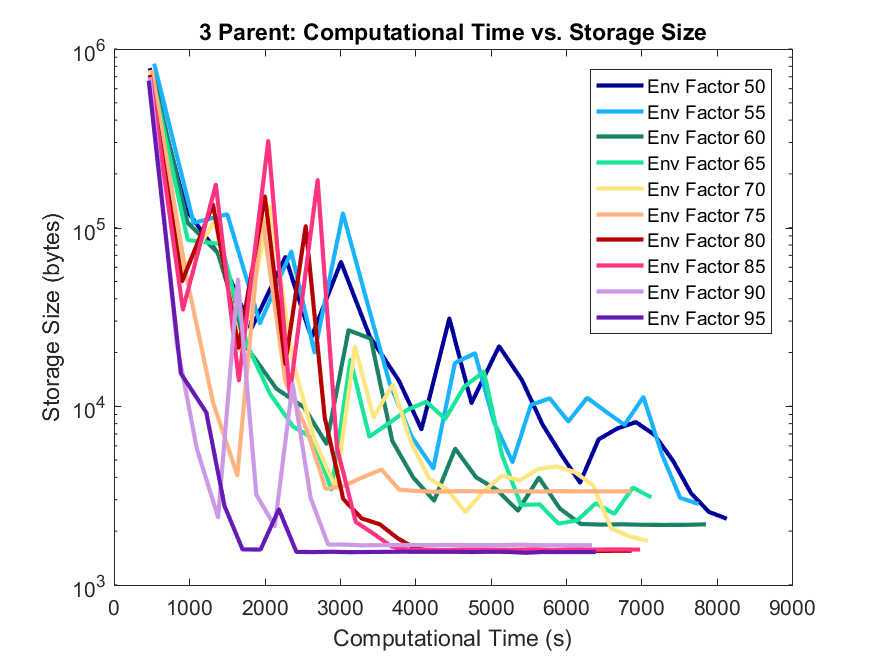}\\\\
\includegraphics[width=.45\textwidth]{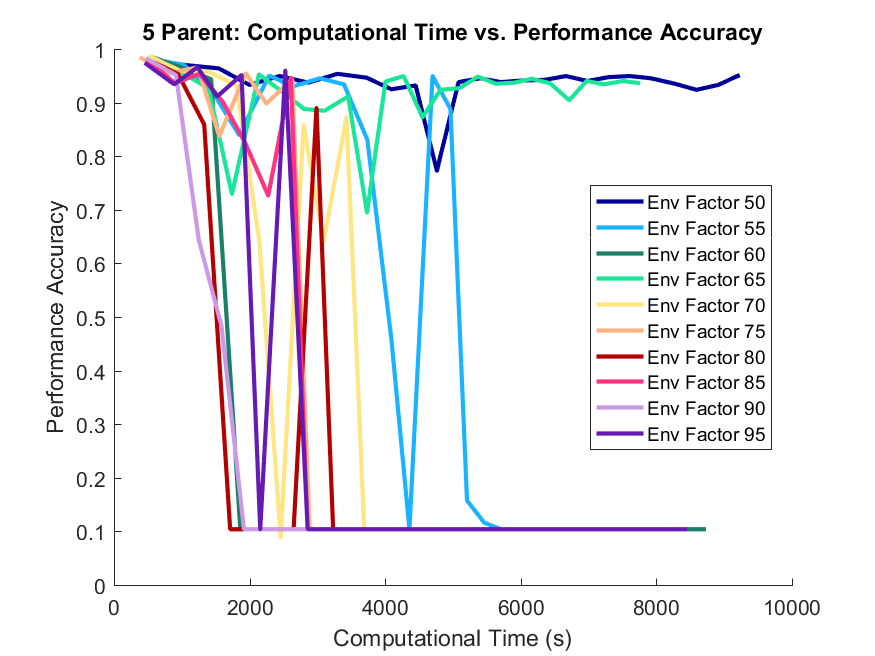}&
\includegraphics[width=.45\textwidth]{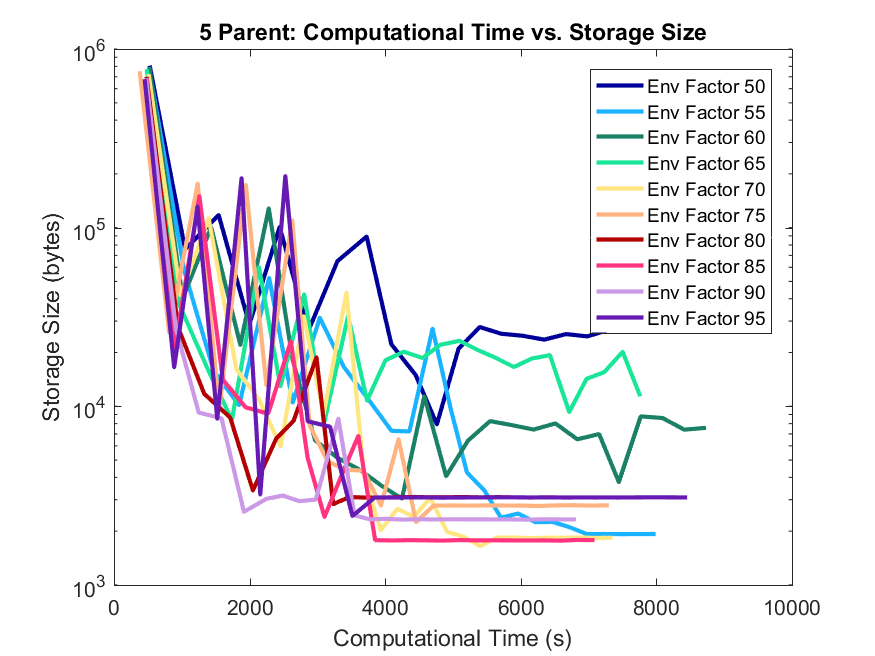}\\\\
\end{tabular}
\caption{Performance accuracy (left) and storage size (right) with respect to computational time for 2-parent evolutionary synthesis (top row), 3-parent evolutionary synthesis (middle row), and 5-parent evolutionary synthesis (bottom row) using various environmental factor models.}
\label{fig_Results235}
\end{figure*}

Figure~\ref{fig_Results1} shows the performance accuracy and storage size for 1-parent evolutionary synthesis using various cluster-level environmental factor models, and shows a clear monotonic trend with respect to the environmental factors. As the environmental factor model decreases, the rate of decrease in storage size of synthesized networks slows accordingly and the rate of performance accuracy loss over generations is similarly slowed. As such, Figure~\ref{fig_Results1} shows that networks synthesized with an environmental factor of 50\% (dark blue) has the most gradual decrease in performance accuracy and storage size, while networks synthesized with an environmental factor of 90\% (light purple) or 95\% (dark purple) have the steepest decrease in performance accuracy and storage size over successive generations.

One aspect to note in Figure~\ref{fig_Results1} is the bottom plateau in performance accuracy at 10\%; this is due to the MNIST dataset~\cite{LeCun1998} itself which consists of 10 classes of individual handwritten digits (i.e., digits 0 to 9), and the 10\% performance accuracy is akin to random guessing. There is similarly a bottom plateau in network storage size that is most easily seen in the trends for environmental factors of 90\% (light purple) and 95\% (dark purple). This plateau corresponds to networks where there is only a single synapse in each convolution layer and, thus, cannot be reduced any further. This network storage size reduction of approximately three orders of magnitude is a result of using the LeNet-5~\cite{LeCun1998_LeNet} architecture as the first generation ancestor network, and there is no inherent limit in network reduction when using the evolutionary deep intelligence approach.

The performance accuracy and storage size for 2-parent, 3-parent, and 5-parent evolutionary synthesis using various cluster-level environmental factor models (Figure~\ref{fig_Results235}) show similar general trends to 1-parent evolutionary synthesis in Figure~\ref{fig_Results1}. There is noticeably more variability in both performance accuracy and storage size as the number of parent networks increases; this is likely a result of only one synthesized network being represented per generation in Figure~\ref{fig_Results235}. Lastly, while there is synaptic competition as imposed by the environmental factor models, the network synthesis process is non-competitive between parent networks (i.e., no selection criteria) and potentially contributes to the variability in performance accuracy and storage size over successive generations.

\begin{figure*}
\centering
\begin{tabular}{cc}
\includegraphics[width=.48\textwidth]{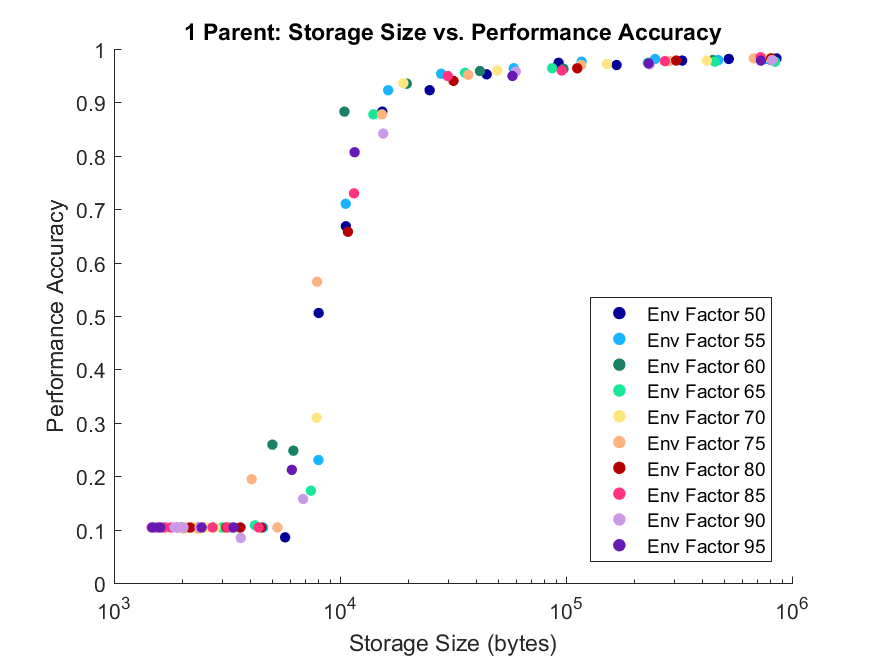}&
\includegraphics[width=.48\textwidth]{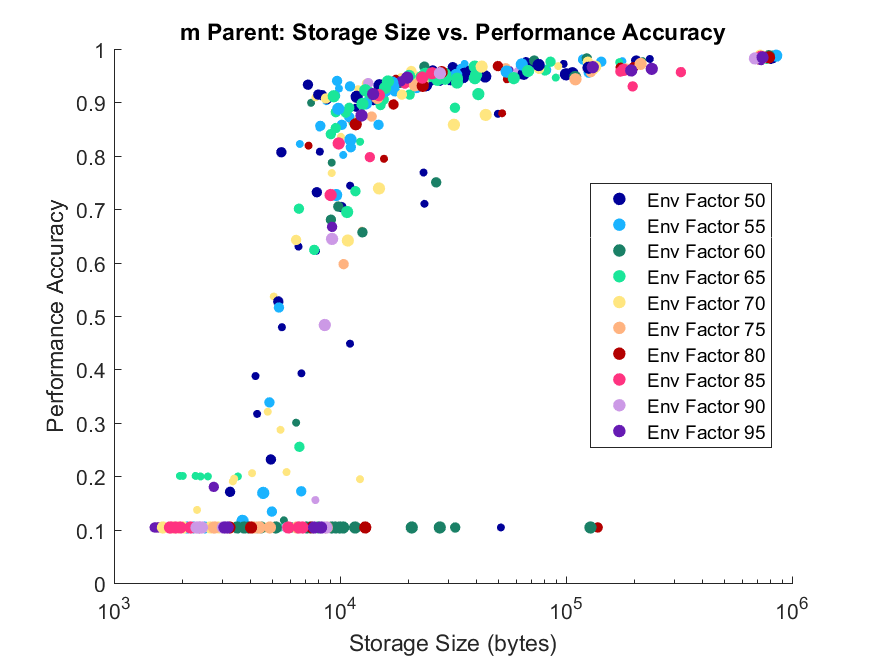}\\\\
\end{tabular}
\caption{Performance accuracy as a function of storage size for 1-parent asexual evolutionary synthesis (left) and $m$-parent sexual evolutionary synthesis (right) using various environmental factor models. 2-parent evolutionary synthesis (smallest points), 3-parent evolutionary synthesis (medium points), and 5-parent evolutionary synthesis (largest points) show noticeably more variation than the 1-parent case.}
\label{fig_ResultsScatter}
\end{figure*}

Figure~\ref{fig_ResultsScatter} shows performance accuracy as a function of storage size for 1-parent asexual evolutionary synthesis (left) and $m$-parent sexual evolutionary synthesis (right) using various cluster-level environmental factor models, where the best synthesized networks are closest to the top left corner, i.e., high performance accuracy and low storage size. In the $m$-parent sexual evolutionary synthesis figure, 2-parent evolutionary synthesis is represented using the smallest points, 3-parent evolutionary synthesis is represented by medium-sized points, and 5-parent evolutionary synthesis is represented using the largest points.

Notice that in 1-parent evolutionary synthesis, most of the networks closest to the top left corner are synthesized using environmental factors of 50\%, 55\%, and 60\%. Similarly for $m$-parent evolutionary synthesis, the best networks (regardless of number of parents) are generally synthesized using the lowest environmental factor models.  Specifically, environmental factor models of 50\% (dark blue), 55\% (light blue), and 60\% (dark green) produced the majority of the synthesized networks closest to the top left corner.

\section{Conclusion}
\label{Conclusion}
In this work, we examined the role external environmental resource models played during the evolutionary synthesis process. To achieve this, we varied the availability of simulated environmental resources (using environmental factor models of 50\% to 95\% at 5\% increments) in the context of $m$-parent evolutionary synthesis for $m = 1, 2, 3, 5$ over successive generations, and analysed the resulting trends in performance accuracy and network storage size.

Overall, a lower environmental factor model resulted in a more gradual loss in performance accuracy and decrease in storage size. This potentially allows for significant reduction in storage size with minimal drop in performance accuracy (e.g., environmental factor 50\% in 5-parent evolutionary synthesis). In addition, the best networks were synthesized using the lowest environmental factor models, i.e., environmental factors of 50\%, 55\%, and 60\%.  However, there is a tradeoff between decreasing the environmental factor model (and, as a result, the rate at which newly synthesized networks drop synapses) and increasing network training time due to a higher number of synapse (i.e., trainable parameters).

Future work includes the investigation of a time-varying environmental factor models (e.g., increasing or decreasing the available resources over generations) to imitate changing environmental conditions, and the incorporation of parent network selection criteria to mimic natural selection in the face of scarce environmental resources. In addition, further analysis of the variability in performance accuracy and storage size over successive generations is necessary, particularly with increasing the number of parent networks during evolutionary synthesis.

\section*{Acknowledgement}
This work was supported by the Natural Sciences and Engineering Research Council of Canada  and Canada Research Chairs Program. The authors also thank Nvidia for the GPU hardware used in this study through the Nvidia Hardware Grant Program.

\bibliographystyle{IEEEtran}

\bibliography{EvoNetResources}

\end{document}